\documentclass[review]{elsarticle} %

\usepackage{amsthm}

\usepackage{graphicx}
\usepackage{amsmath,amssymb,amsfonts}
\usepackage{algorithmic}
\usepackage{textcomp}
\usepackage{xcolor}
\usepackage{multicol}
\usepackage{multirow}
\usepackage{makecell}
\usepackage{adjustbox}
\usepackage{tabularx}
\usepackage{float}
\usepackage{kotex}
\usepackage{csquotes}
\usepackage{color, colortbl}
\usepackage{arydshln}

\usepackage[final]{microtype}

\definecolor{Gray}{gray}{0.95}
\definecolor{Yellow}{RGB}{255, 255, 153}
\def\BibTeX{{\rm B\kern-.05em{\sc i\kern-.025em b}\kern-.08em
    T\kern-.1667em\lower.7ex\hbox{E}\kern-.125emX}}
    
\newcommand*\colourcheck[1]{%
  \expandafter\newcommand\csname #1check\endcsname{\textcolor{#1}{\ding{52}}}%
} 
\usepackage{soul}
\colorlet{tdcolor}{yellow!35}
\sethlcolor{tdcolor}

\journal{Pattern Recognition}

\begin{document}

\begin{frontmatter}



\title{\bf Enhancing Spatio-Temporal Zero-shot Action Recognition with Language-driven Description Attributes}

\author[inst1]{Yehna Kim}
\author[inst1]{Young-Eun Kim}
\author[inst1]{Seong-Whan Lee\corref{cor1}}

\affiliation[inst1]{organization={Department of Artificial Intelligence, Korea University},
            addressline={Anam-dong, Seongbuk-gu}, 
            city={Seoul},
            postcode={02841}, 
            country={Korea}}


\cortext[cor1]{Corresponding author \ead{sw.lee@korea.ac.kr}}





\begin{abstract}
Vision-Language Models (VLMs) have demonstrated impressive capabilities in zero-shot action recognition by learning to associate video embeddings with class embeddings. However, a significant challenge arises when relying solely on action classes to provide semantic context, particularly due to the presence of multi-semantic words, which can introduce ambiguity in understanding the intended concepts of actions. To address this issue, we propose an innovative approach that harnesses web-crawled descriptions, leveraging a large-language model to extract relevant keywords. This method reduces the need for human annotators and eliminates the laborious manual process of attribute data creation. Additionally, we introduce a spatio-temporal interaction module designed to focus on objects and action units, facilitating alignment between description attributes and video content. In our zero-shot experiments, our model achieves impressive results, attaining accuracies of 81.0\%, 53.1\%, and 68.9\% on UCF-101, HMDB-51, and Kinetics-600, respectively, underscoring the model's adaptability and effectiveness across various downstream tasks.
\end{abstract}




\end{frontmatter}

\section{Introduction}
Vision-language models (VLMs)\cite{clip, flamingo}, trained extensively on diverse datasets containing image-text pairs using contrastive learning techniques, have showcased impressive capabilities across various tasks\cite{prj-2, prj-3}. Through their training process, VLMs effectively align images and corresponding text descriptions into a unified latent space, enabling them to excel not only in closed-set scenarios but also in zero-shot transfer settings. Their remarkable performance spans a wide array of downstream vision-related tasks. Moreover, recent efforts\cite{text4vis, maxi, asu} have extended the application of VLMs to video data, aiming to capitalize on their robust zero-shot performance. These endeavors focus on learning the intricate relationship between video embeddings and class embeddings, leading to significant enhancements in zero-shot task performance.

\begin{figure}[t]
    \centering
    \includegraphics[width=1.0\linewidth]{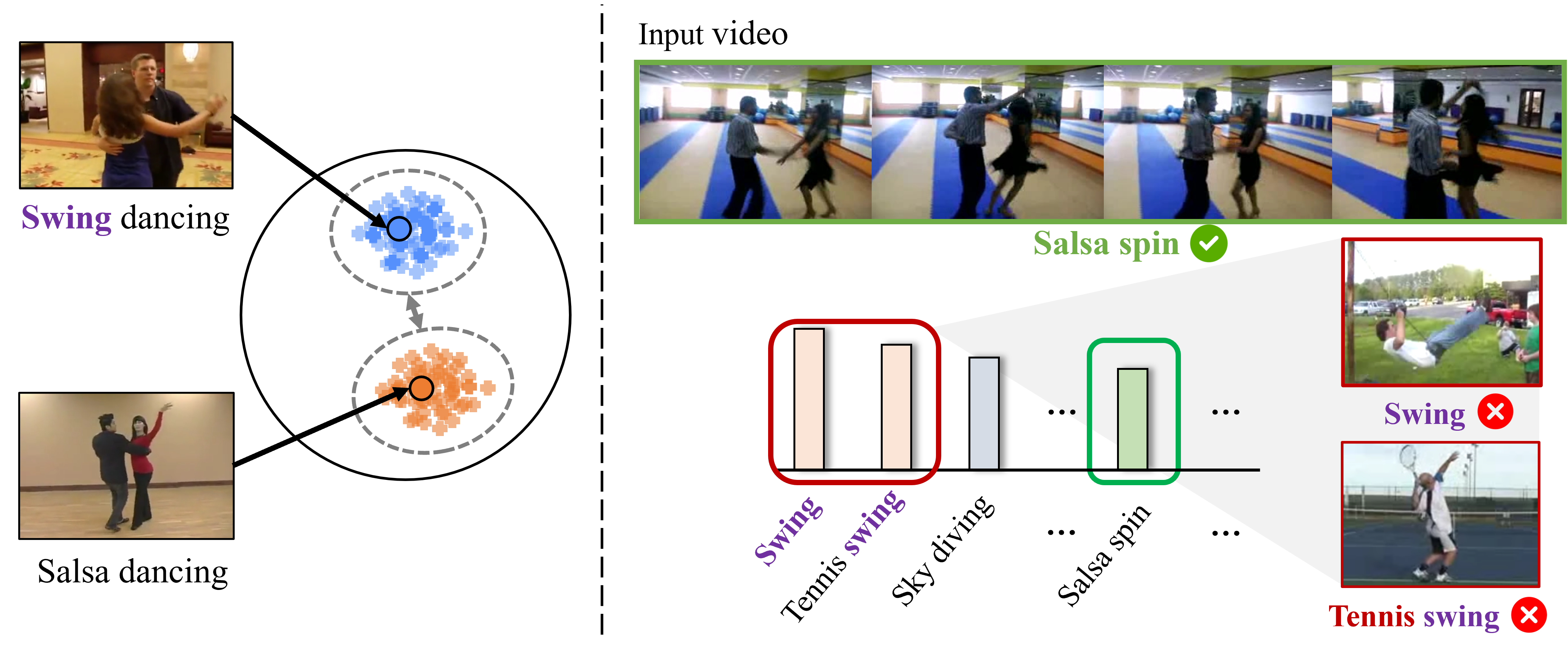}
    \caption{Example of misclassified data due to the ambiguity of action classes. The model incorrectly infers \textit{\enquote{salsa spin}} as \textit{\enquote{swing}} or \textit{\enquote{tennis swing}}, due to multi-semantic word \textit{swing}. This error illustrates the need for additional semantic information beyond action class labels.}
    \label{fig:issue}
\end{figure}

Many zero-shot action recognition models rely heavily on action classes as the primary source of semantic information for their general representation. However, an inherent issue arises when action classes are used in this capacity. The crux of the problem lies in the presence of multi-semantic words within these action classes—words that share identical spellings but possess distinct meanings in various contexts. Consequently, during the training process, the interpretation of different actions and their contextual significance becomes muddled by these multi-semantic words, complicating the task of accurately capturing the intended concept behind each action.

To illustrate, consider the scenario depicted in Fig. \ref{fig:issue}, wherein a zero-shot learning process is conducted on the Kinetics-400 dataset \cite{kinetics400} followed by inference on the UCF-101 dataset \cite{ucf101} without further training. The model frequently misclassifies videos depicting \enquote{salsa spin} as either \enquote{swing} or \enquote{tennis swing}. This misclassification stems from a confusion regarding the meaning of the word \enquote{swing}. During training, the model learns from \enquote{swing dance} videos and associates them closely with similar \enquote{salsa dance} videos. However, when inferring on unseen datasets like UCF-101, labels containing the word \enquote{swing} may refer to unrelated concepts such as a \enquote{swinging seat} or \enquote{tennis swing}. Consequently, when presented with a \enquote{salsa spin} video, the model erroneously classifies it as \enquote{swing} or \enquote{tennis swing}. This underscores the necessity for supplementary semantic information during the training phase to facilitate the acquisition of nuanced representations underlying action classes.

\begin{figure}    
    \centering
    \includegraphics[width=1\linewidth]{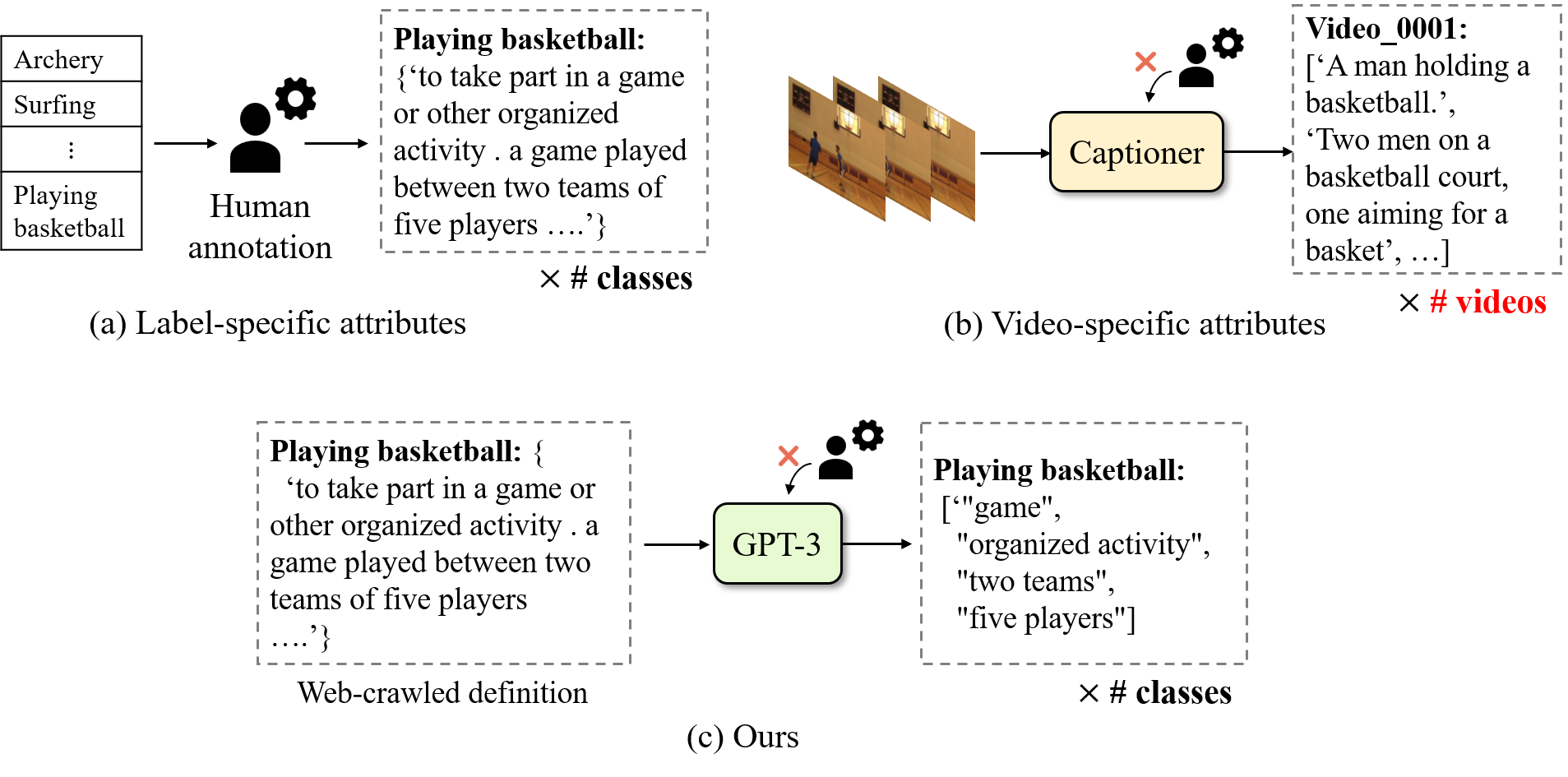}
    \caption{Illustration of the difference between label-specific attributes (a), video-specific attributes (b) and ours (c). Our approach eliminates the manual annotation process and achieves zero-shot performance using only label-specific attributes.}
    \label{fig:attributes}
\end{figure}

Recent research efforts have embraced the integration of attribute data alongside action classes to enhance the breadth of representations \cite{asu, bike, maxi}. This integration typically falls into two categories: label-specific attributes \cite{asu, victr} and video-specific attributes \cite{bike, maxi} (depicted in Fig. \ref{fig:attributes}). Label-specific attributes are crafted through the utilization of web-crawled dictionary definitions and the manual curation of attribute data, enriching the semantic understanding associated with each class label. While these attributes provide precise semantic details regarding the action classes, the manual construction process can be resource-intensive.

On the other hand, video-specific attributes extract additional semantic cues directly from video content using methodologies such as object detectors \cite{st-clip} or captioners \cite{maxi}. This approach streamlines the process by reducing the manual labor required in the label-specific approach and facilitates the extraction of instance-level attributes within frames. However, replicating video-specific attributes on unseen datasets necessitates the creation of attributes for each video within the dataset. Moreover, it's important to note that video-specific attributes may encompass information unrelated to the actual action class, presenting a challenge in maintaining relevance and coherence within the attribute data.

We introduce a novel approach for generating label-specific descriptions using large-language models, aimed at circumventing the labor-intensive manual annotation process. 
Our method leverages web-crawled descriptions that are contextually relevant to each action class and extracts keywords from these descriptions using a large-language model, specifically GPT-3 \cite{gpt3}. This ensures that the attributes selected by the large-language model are pertinent to each action class, thereby reducing reliance on human annotators. By employing label-specific attributes rather than video-specific attributes, our approach not only reduces the cost of manual annotation but also achieves commendable zero-shot performance.

Furthermore, we propose a spatio-temporal interaction module designed to align description attributes with video frames effectively. These attributes encapsulate information about objects or action units, corresponding to specific segments of the video. The spatio-temporal interaction module facilitates the model in comprehensively understanding and capturing the detailed concepts of attributes by mapping spatial and temporal information on a fine-grained basis.

To validate the transferability of our model to downstream tasks, we conduct experiments on standard zero-shot video recognition datasets, including Kinetics-400 \cite{kinetics400}, Kinetics-600 \cite{kinetics600}, HMDB-51 \cite{hmdb}, and UCF-101 \cite{ucf101}. Our findings demonstrate significant performance improvements across zero-shot, few-shot, and fully-supervised recognition scenarios. Specifically, in the zero-shot setting, our model achieves an accuracy of 81.0\% on UCF-101 and 53.1\% on HMDB-51. The primary contributions of our study are outlined as follows:

\begin{itemize}
\item Our proposed approach leverages action class descriptions by extracting more meaningful words using a large-language model, thereby achieving cost-effective zero-shot performance without the need for video-specific attributes.
\item We introduce a spatio-temporal interaction module that enhances the alignment between description attributes and video embeddings by incorporating spatial and temporal information on a fine-grained basis.
\item Experimental results demonstrate the transferability of our model across zero-shot, few-shot, and fully-supervised recognition scenarios, achieving zero-shot accuracies of 81.0\% on UCF-101 and 53.1\% on HMDB-51.
\vspace{0.4cm}
\end{itemize}

\section{Related Work}

\subsection{Diverse Approaches to Action Recognition}
While action recognition on RGB videos (e.g., \cite{prml-1,prml-2,prml-3,prml-6,prml-7, prj-1}) has garnered significant attention, several works have explored alternative modalities or scenarios such as skeleton-based action understanding or group activity recognition. For instance, Xu \textit{et al.} \cite{xu2023} 
proposed an auxiliary-task-driven transformer for robust 3D skeleton motion prediction, and Huang \textit{et al.} \cite{huang2023higcin} extended graph-based methods with a hierarchical cross-inference network for complex group scenarios. Similarly, Zhang \textit{et al.} \cite{zhang2023multigran} proposed a multi-granularity anchor-contrastive approach to address semi-supervised \emph{skeleton-based} action recognition. 

These approaches underscore the breadth of research in action understanding, ranging from single-person skeleton estimation to multi-person group coordination. However, they typically rely on specialized datasets (e.g., NTU RGB+D for skeletons, Volleyball dataset for group activity), which differ from large-scale RGB benchmarks like Kinetics or UCF-101 that we target in this work.

\subsection{Action Recognition with VLMs}
VLMs have demonstrated remarkable progress across various tasks, prompting their extension into the realm of video action recognition, a burgeoning trend in recent research. For instance, ActionCLIP \cite{actionclip} introduced a novel visual prompt approach, integrating a transformer post-network prompt to facilitate temporal interactions. Similarly, X-CLIP \cite{x-clip} proposed a cross-frame attention mechanism, leveraging video embeddings to enhance text prompts, thereby refining text embeddings. Another notable contribution comes from Text4Vis \cite{text4vis}, which investigates the role of linear classifiers and introduces alternative knowledge sources to replace the pre-trained model's classifier. In contrast to existing methodologies that heavily rely on action label classes for semantic cues, our approach diverges by incorporating descriptive attributes. This novel incorporation enhances the efficacy of pre-trained VLMs, empowering the model to capture more intricate and nuanced video representations. 

\subsection{Zero-shot Action Recognition}
Zero-shot action recognition seeks to develop a comprehensive understanding of videos, enabling inference on unseen data not encountered during training. However, inherent limitations exist within action classes themselves due to their varying semantic interpretations across different contexts. Prior research efforts have followed two complementary streams: (i) attribute integration as supplemental semantics, categorized into label-specific and video-specific types \cite{asu,victr,maxi,bike} and (ii) representation-centric approaches that decouple appearance/motion or spatial/temporal factors for sample-efficient few- and zero-shot recognition \cite{xing2021,su2021}. Our approach adopts an attribute-centric perspective, emphasizing label-specific descriptions rather than representation decoupling.

Label-specific attributes are derived from individual class labels. For instance, ASU \cite{asu} organizes semantic unit attributes into objects, scenes, body parts, and motion, necessitating manual construction and categorization. Conversely, video-specific attributes extract additional semantic insights from videos using methods such as object detection or captioning. BIKE \cite{bike}, for example, utilizes video-associated attributes by selecting the top-k relevant classes among action classes for each video. However, attributes generated through this approach may contain irrelevant information.

Recently, VideoPrompter\cite{videoprompter} proposed a GPT-based zero-shot video understanding approach that shares some conceptual overlap with our language-driven attribute extraction, but our work differs by focusing on spatio-temporal fusion centered on extracted keywords. Meanwhile, GPT4Vis \cite{gpt4vis} merely expands upon category names, contrasting our strategy of deriving label-specific attributes via carefully selected keywords for more precise zero-shot recognition.

In our approach, we introduce label-specific description attributes, avoiding manual intervention while offering instance-level granularity to enhance semantic richness in video representation.

\section{Method}
We propose a spatio-temporal interaction mechanism to refine the alignment between description attributes at a fine-grained level. Initially, we generate description attributes (Section \ref{sec:descattr}), followed by their incorporation into the network architecture. The architecture encompasses spatial and temporal interaction modules, facilitating precise alignment between attribute word embeddings and frame embeddings. This approach offers an intricate comprehension of the spatio-temporal dynamics within videos.

\subsection{Preliminary}
We provide an overview of the standard procedure for video recognition using vision-language models (VLMs), which effectively integrate visual and textual data. As our foundational architecture, we utilize CLIP \cite{clip}, renowned for its consistent performance across diverse downstream applications. CLIP comprises a visual encoder $f(\cdot|\theta_{v})$ and a text encoder $g(\cdot|\phi_{c})$. Our objective is to embed each video and class label using the CLIP encoder, ensuring that their representations are proximate in latent space. Given a dataset of videos $\mathcal{V}=\{\boldsymbol{v}_i\}_{i=1}^{n}$ (comprising $T$ image frames, i.e., $\boldsymbol{v}_i = \{v_i^1 , v_i^2 , \cdots , v_i^T\}$) and action classes $\mathcal{C}=\{{c}_{j}\}_{j=1}^{m}$, we extract the embeddings with CLIP as follows:
\begin{equation}
    \mathbf{z}_{{v}_i}^t = f\left({v}_i^t | \theta_{v}\right), \; \mathbf{z}_{c_j} = g\left(c_j |\phi_{c}\right),
\end{equation}
{where $t$ is the frame index.}
Some VLMs-based methods \cite{finetunedCLIP,maxi} derive video embeddings by averaging the frame embeddings, i.e., $\mathbf{z}_{\boldsymbol{v}_i}=\sum_t f\left(v_i^t | \theta_v\right) / T$.
The model is then trained using a similarity score,  $s(\mathbf{z}_{{v}_i}, \mathbf{z}_{c_j})$, which assesses the resemblance between a video $\boldsymbol{v}_i$ and a class label $c_j$.

{However, this naive averaging discards crucial temporal dependencies and spatial details, leading to a loss of fine-grained contextual information. Instead, we propose a Spatial-Temporal Interaction Module, which dynamically refines feature representations per frame while integrating relevant Descriptive Attributes for enhanced semantic grounding.} We maintain the same notation throughout this paper.

\begin{figure}[t]
    \centering
    \includegraphics[width=1.\linewidth]{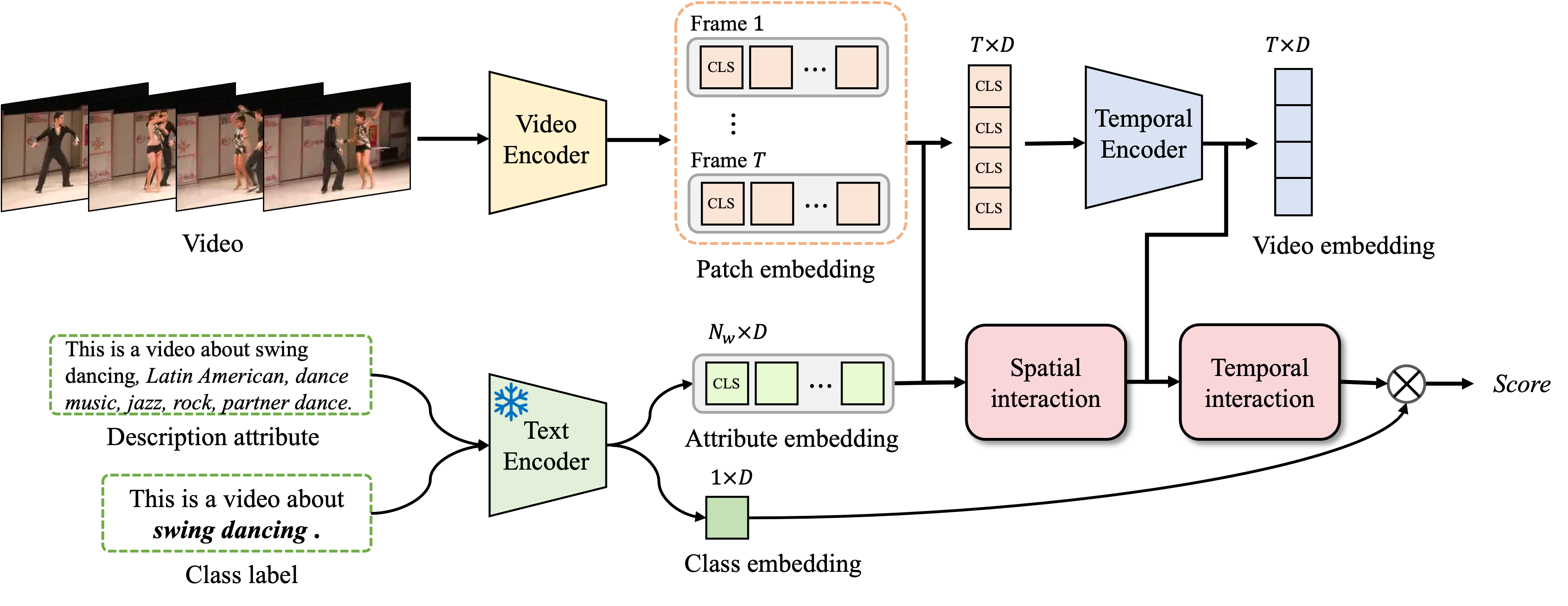}
    \caption{An overview of our model framework. The architecture consists of spatial interaction and temporal interaction modules.}
\end{figure}

\subsection{Descriptive Attributes with text expansion} 
\label{sec:descattr}
{We introduce a novel approach for generating label-specific descriptions, Descriptive Attributes (DAs), using large-language models, aimed at circumventing the labor-intensive manual annotation process. Our method leverages web-crawled descriptions that are contextually relevant to each action class and extracts keywords from these descriptions using a large-language model, specifically GPT-3} \cite{gpt3}. 

{In practice, we extract description attributes based on the elaborative description used in ER} \cite{erzsar}. {This description is refined to the smallest set of sentences by crawling action class definitions from Wikipedia and dictionaries. Then, we pass these descriptions to GPT-3 using a custom prompt, such as: ``Extract 5-10 essential keywords from \{\textit{description}\} that best describe the action \{\textit{action name}\} in the paragraph. Focus on objects, motions, and contexts related to the action." We use GPT-3 with $temperature = 0.7$ and $max\_tokens = 256$ to encourage diverse, concise outputs. We filter out high-frequency stopwords (like `the', `and', etc.) and duplicates. Finally, we retain the top-N (e.g., $N=8$) candidate keywords, which collectively form the descriptive attributes for the action class.}

The selected keywords are aggregated and we define a DA set for action $c_j$ as, 
\begin{equation}
\operatorname{DA}(c_j) = \{w_j^1, \cdots, w_j^{N_a}\},
\end{equation}
where $N_a$ represents the number of keywords used for descriptive attributes. 
Following the \cite{erzsar}, we form a composite token by concatenating the class label and its corresponding DAs as follows:
\begin{equation}
    c_j^{DA}  = \left[c_j; w_j^1; \cdots; w_j^{N_a}\right].
\end{equation} 
Since the action class $c_j$ is extended with extra information using DA, $c_j^{DA}$ will possess richer and more accurate details about $c_j$, thereby avoiding the ambiguity issue with action classes discussed in Figure \ref{fig:issue}.

This concatenated token sequence $c_j^{DA}$ is then prompted with a predefined text template, \enquote{This is a video about \{\}.} generating a cohesive attribute sentence. To extract the text embedding, the sentence undergoes encoding with the CLIP text encoder to producing an attribute embedding as,

\begin{equation}
    \mathbf{z}_{c_j^{DA}} = g(c_j^{DA}; \phi_{c}),
\end{equation}
where $\mathbf{z}_{c_j^{DA}} = \left[\mathbf z_{c_{j,cls}}; a_{j,1}; \cdots ;a_{j,N_w}\right]$, $z_{c_{j,cls}}$ is the class embedding, and $a_{j,N_w}$ is the attribute embedding with \(N_w\) indicating the number of word tokens in the attribute sentence. Here, we define the attribute embeddings for the entire word tokens $N_w$ as $\mathbf a_{j} \in \mathbb{R}^{N_w \times D}$ and \(D\) signifying the dimensionality of the text embeddings.

\begin{figure*}[t]
    \centering
    \includegraphics[width=1\linewidth]{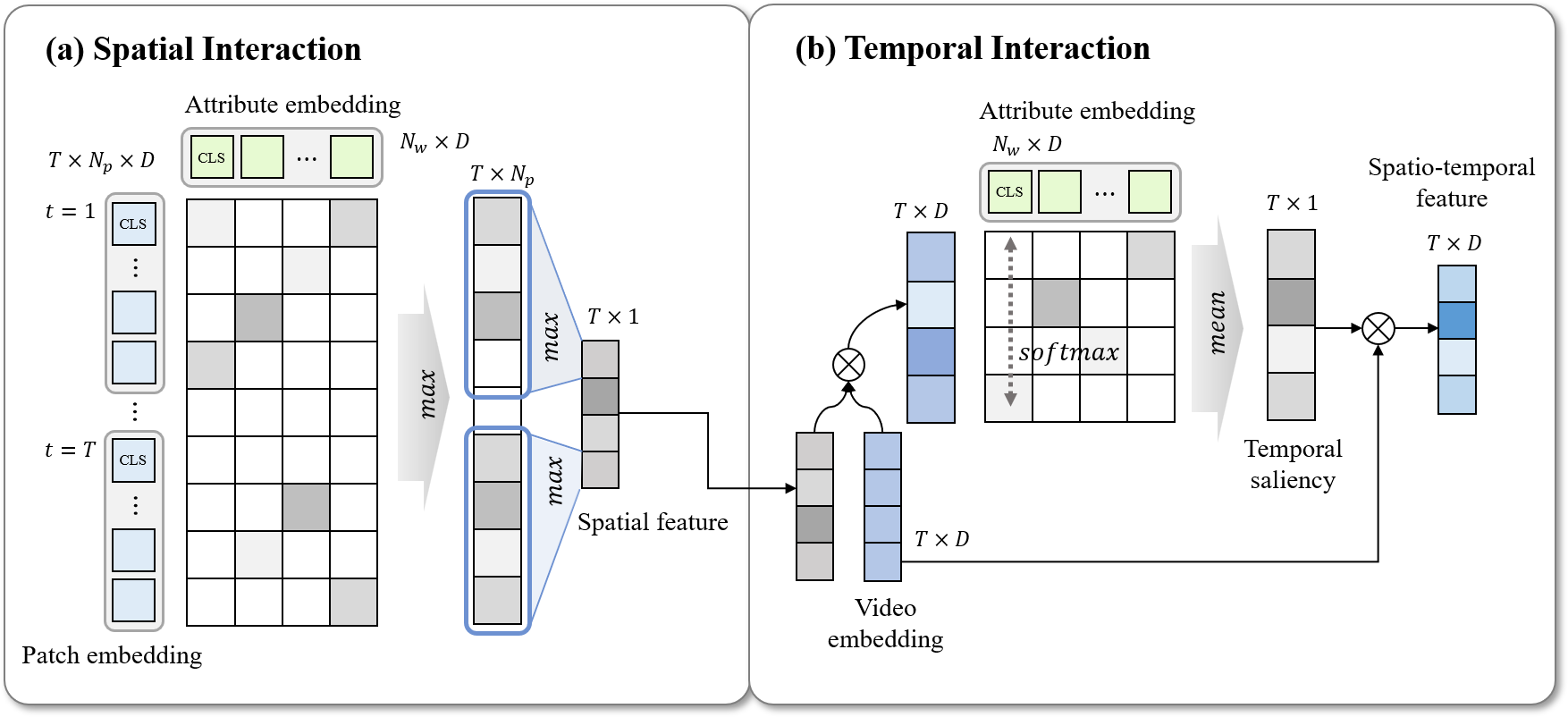}
    \caption{Details of spatio-temporal interaction. \textbf{(a) Spatial interaction:} Patch embeddings from each frame $(T \times N_p \times D)$ and attribute word embeddings $(N_w \times D)$ are projected to a common space. Patch-word similarities are computed and max-pooled across words (per patch) and then max-pooled across patches (per frame) to yield spatial features $f_{sp} \in \mathbb{R}^{T \times 1}$. \textbf{(b) Temporal interaction: } Word-frame similarities are softmax-normalized along time and averaged across word to produce temporal saliency $\boldsymbol{S}_{temp} \in \mathbb{R}^{T \times 1}$; these weights re-scale the video embedding to obtain the final spatio-temporal feature $(T\times D)$.}
    \label{fig:framework}
\end{figure*}

\subsection{Spatial-Temporal Interaction} 
\label{sec:spatial}
In this section, we present the Spatial-Temporal Interaction (STI) module, designed to enable interaction between visual features and DA-enhanced text features. For this purpose, we develop spatial and temporal interaction modules, which generate frame-by-frame refined visual embeddings and refined text embeddings, respectively.

\paragraph{Spatial Interaction}
The objective of the spatial interaction module is to calculate detailed correlations and extract prominent visual features, as illustrated in Figure. \ref{fig:framework}(a). 
Given a video input  consisting of uniformly sampled $T$ image frames, $\boldsymbol{v}_i = [v_i^1 ; v_i^2 ; \cdots ; v_i^T]$, we embed each frame using the CLIP image encoder to generate a sequential feature:
\begin{equation}
    \mathbf{z}_{v_i^t} = f(v_i^t; \theta_v), \quad t=1,\cdots,T,
\end{equation}
where $\mathbf{z}_{v_i^t} = [\mathbf z_{i,cls}^t; p_{i,1}^t; \cdots p_{i,N_p}^t]$, with $N_p$ indicating the number of patches. We define the patch embeddings for the entire $T$ frame as $\mathbf p_{i} \in \mathbb{R}^{T \times N_p \times D}$.

Next, We perform patch embedding projection and word embedding projection to emphasize meaningful patch and word tokens. This involves applying a linear layer with ReLU activation function, utilizing weights $\mathbf{W}_p, \textbf{W}_t \in \mathbb{R}^{D \times D}$, where $D$ denotes the dimensions of video and word embedding. The projected patch embedding $\boldsymbol{p}_{\mathrm{proj}}$ and projected word embedding $\boldsymbol{a}_{\mathrm{proj}}$ are defined as:

\begin{equation}
    \tilde{\mathbf p}_{i} = ReLU(p_i \cdot \mathbf{W}_p), \quad
    \tilde{\mathbf{a}}_{j} = ReLU(\mathbf a_{j} \cdot \mathbf{W}_w,)
\end{equation} 
where $\tilde{\mathbf p}_{i} \in \mathbb{R}^{N_p \times T \times D}$ and $\tilde{\mathbf{a}}_{j} \in \mathbb{R}^{N_w \times D}$.

We then calculate the maximum similarity of weighted patches and words within each frame, associating the most similar words to each projected patch, thereby reflecting cross-modal fine-grained matching. Finally, we compute the dot product of the patch-word matching scores $\boldsymbol{S}_{sp}$ with the video embedding $v_i$ to obtain the final spatial features $\boldsymbol{f}_{sp}$. The patch-wise spatial interaction process is defined as:
\begin{gather}
    \boldsymbol{S}_{sp} = \max_{l=1}^{N_w} \max _{k=1}^{N_p}
    \left[\tilde{\mathbf p}_{i} \cdot \tilde{\mathbf{a}}_{j}^\top \right]_{kl}, \\
    \boldsymbol{f}_{sp,i} = \boldsymbol{S}_{sp} \cdot \boldsymbol v_i,
\end{gather}
where $\boldsymbol{S}_{sp} \in \mathbb{R}^{T}$, $\boldsymbol v_i \in \mathbb{R}^{T \times D}$ and $\boldsymbol{f}_{sp,i} \in \mathbb{R}^{T \times D}$.

\paragraph{Temporal Interaction}
In the temporal interaction module, we aim to estimate temporal saliency for video feature aggregation, as depicted in Fig. \ref{fig:framework}(b). Drawing inspiration from video concept spotting techniques \cite{bike}, we leverage the spatial features and attribute word embeddings to capture temporal saliency for video feature aggregation. By utilizing attribute word embeddings as queries, we achieve a more granular saliency estimation at the word-to-frame level.

To compute temporal saliency, we calculate the similarity between each word and each frame, followed by a softmax operation to normalize these similarities for each frame. Aggregating these normalized similarities across various words for a specific frame, we obtain the overall saliency $\boldsymbol{S}_{temp} \in \mathbb{R}^{T}$ at the frame level, defined as:

\begin{equation}
\boldsymbol{S}_{temp}= \frac{1}{N_w} \sum_{n=1}^{N_w} \frac {\exp ((\boldsymbol{f}_{sp,i}^t)^\mathsf{T} \tilde{\mathbf{a}}_{j}^n/\tau)}{\sum_{t=1}^T \exp ((\boldsymbol{f}_{sp,i}^t)^\mathsf {T} \tilde{\mathbf{a}}_{j}^n/\tau )}, 
\label {eq:softmax}
\end{equation}
where $\boldsymbol{f}_{sp,i}^t$ represents the $t$-th frame of the spatial feature $\boldsymbol{f}_{sp,i}$, and $\tau$ denotes the temperature parameter of the softmax function.

Subsequently, we utilize the temporal saliency to aggregate these frame embeddings as follows:
\begin{equation}
\boldsymbol{f}_{st,i} = \sum_{t=1}^T v_i^t \boldsymbol{S}_{temp},
\end{equation}
where $\boldsymbol{f}_{st,i} \in \mathbb{R}^D$ represents the final video representation enhanced by spatio-temporal interaction for the video $\boldsymbol{v}_i$.

\subsection{Training Objectives}
We leverage the textual features of action labels to guide the refinement of video representations. The parameters of the video encoder are initialized with weights from a pre-trained Vision-Language Model (VLM), while the parameters of the pre-trained text encoder remain fixed.

During the training process, our aim is to ensure that the final video embedding $\boldsymbol{f}_{st}$ and the class embedding $z_{i,cls}$ exhibit similarity when they correspond to related concepts, and dissimilarity otherwise. Let $\mathcal{C}$ denote the set of $K$ categories indexed by $y_i \in [1, K]$, where $y_i$ represents the label indicating the index of the category in the dataset.

Following the bidirectional learning objective outlined in \cite{actionclip}, we employ symmetric cross-entropy loss to maximize similarity between matched video representations and class embeddings, while minimizing similarity for other pairs:

\begin{gather}
    \mathcal{L}_{V2C} = -\frac 1 B \sum_i^B \frac{1}{|\mathcal{K}(i)|} \sum_{k \in \mathcal{K}(i)}\log\frac{\exp(s(z_{c_{i,cls}}, \boldsymbol{f}_{st, k})/\tau)}{\sum_j^B \exp(s(z_{c_{i,cls}}, \boldsymbol{f}_{st,j})/\tau)}, \\
    \mathcal{L}_{C2V} = -\frac 1 B \sum_i^B \frac{1}{|\mathcal{K}(i)|} \sum_{k \in \mathcal{K}(i)}\log\frac{\exp(s(z_{c_{k,cls}}, \boldsymbol{f}_{st,i})/\tau)}{\sum_j^B \exp(s(z_{c_{j,cls}}, \boldsymbol{f}_{st,i})/\tau)}.
\end{gather}
where $k \in \mathcal{K}(i) = \{k|k \in [1,B], c_k=c_i\}$, and $s(\cdot,\cdot)$ represents the cosine similarity.

The total loss is the average of $\mathcal{L}_{V2C}$ and $\mathcal{L}_{C2V}$:
\begin{gather}
    \mathcal{L} = \frac 1 2 (\mathcal{L}_{V2C} + \mathcal{L}_{C2V}).
\end{gather}

At inference, the trained video encoder generates embeddings for unseen videos, which are compared with class embeddings from the text encoder. The video is then classified based on the highest similarity score.


\section{Experiments}
\subsection{Setups}
\subsubsection{Datasets}
We perform experiments on four distinct video datasets: Kinetics-400\cite{kinetics400}, Kinetics-600\cite{kinetics600}, UCF-101\cite{ucf101}, and HMDB-51\cite{hmdb}. Kinetics-400 is a widely used video dataset comprising 400 classes, while Kinetics-600 extends this with 600 classes. UCF-101 contains 13,320 video clips distributed across 101 classes. HMDB-51 consists of 7,000 videos categorized into 51 classes and offers three distinct test data splits. Our experiments encompass fully-supervised recognition on Kinetics-400, zero-shot recognition on UCF-101, HMDB-51, and Kinetics-600, as well as few-shot recognition on UCF-101 and HMDB-51.

\subsubsection{Implementation details} 
{We initialize our video encoder with CLIP's ViT-B/16 model, loading the pre-trained weights $\theta_v$. We fine-tune only the final 6 transformer blocks (out of 12), while the first 6 remain frozen to preserve the original CLIP representations. We train using the AdamW optimizer with a base learning rate of 5e-5 for 30 epochs on the Kinetics-400 dataset (batch size 64, weight decay 0.05). For the text encoder, we keep $\phi_c$ fixed throughout training, allowing the newly introduced parameters ($\mathbf W_p$, $\mathbf W_w,$ and the spatial-temporal modules) to learn to align with the frozen text space.}

{When building descriptive attributes, we set $N_a = 8$ as the default number of keywords extracted from LLM. We tested $N_a \in \{2, 4, 8, 16\}$ and found 8 to offer the best trade-off between performance and computational overhead. Training and evaluation are conducted on 8 A100 GPUs. Our code is implemented in PyTorch 1.13.}

Experiment results are averaged across the three splits, and we report both the top-1 accuracy and standard deviation, in line with \cite{x-clip}. In the case of Kinetics-600, we randomly select 160 categories from the total of 220 new categories for each of the three splits.
In the few-shot scenario, the training set is constructed by randomly sampling 2, 4, 8, and 16 videos from each class. We fine-tune the model on the few-shot dataset for 50 epochs. Evaluation is performed on the first split of the test set, and results are reported based on single-view inference.
For the fully-supervised setting, we employ ViT-L/14 for evaluation on Kinetics-400. Videos are sampled with 8, 16, and 32 frames, and multi-view inference is conducted using 3 spatial crops and 4 temporal clips.

\begin{table*}[t]
\caption{Comparison of zero-shot action recognition performance. Results represent the mean and standard deviation across three validation splits. All models are trained on Kinetics-400. The highest performing model is indicated in \textbf{bold}.}
\centering
\label{tab:zero-shot}
\resizebox{1\linewidth}{!}{
    \begin{tabular}{lccccc}
    \Xhline{2\arrayrulewidth}
    \noalign{\vspace{3pt}}
    \textbf{Method} & \textbf{Encoder} & \textbf{Frames} & \textbf{UCF-101} & \textbf{HMDB-51} & \textbf{Kinetics-600}\\
    
    \noalign{\vspace{1pt}}
    \Xhline{2\arrayrulewidth}
    \multicolumn{6}{l}{\textit{Methods w/o vision-language pre-training}} \\
    \noalign{\vspace{1pt}}

    ER-ZSAR\cite{erzsar} & TSM & 16 & $51.8\pm2.9$ & $35.3\pm4.6$ & $42.1\pm1.4$\\
    JigsawNet\cite{jigsawnet} & R(2+1)D & 16 & $56.0\pm3.1$ & $38.7\pm3.7$ & - \\
    
    \noalign{\vspace{2pt}}
    \hline
    \multicolumn{6}{l}{\textit{Methods w/ vision-language pre-training (ViT-B/16)}}\\
    \noalign{\vspace{1pt}}
    
    X-CLIP\cite{x-clip} & ViT-B/16 & 32 & $72.0\pm2.3$ & $44.6\pm5.2$ & $65.2\pm0.4$\\
    VicTR\cite{victr} & ViT-B/16 & 32 & $72.4\pm0.3$ & $51.0\pm1.3$ & - \\ 
    ASU-B/16\cite{asu} & ViT-B/16 & 8 & $75.0\pm3.7$ & $48.1\pm2.8$ & $67.6\pm0.2$\\
    MAXI\cite{maxi} & ViT-B/16 & 16 & $78.2\pm0.8$ & $52.3\pm0.7$ & $71.5\pm0.8$\\   

    \noalign{\vspace{1pt}}
    \rowcolor{Gray} 
    \textbf{Ours} & ViT-B/16 & 16 & $\mathbf{78.9\pm0.9}$ & $\mathbf{52.7\pm0.8}$ & $\mathbf{72.0\pm0.5}$\\

    \hline
    \multicolumn{6}{l}{\textit{Methods w/ vision-language pre-training (ViT-L/14)}}\\
    \noalign{\vspace{1pt}}
    Text4Vis\cite{text4vis} & ViT-L/14 & 16 & 79.6 & 49.8 & $68.9\pm1.0$\\
    BIKE\cite{bike} & ViT-L/14 & 8 & 80.8 & 52.8 & $68.5\pm1.2$\\
    
    \noalign{\vspace{1pt}}
    \rowcolor{Gray}\textbf{Ours} & ViT-L/14 & 8 & $\mathbf{81.0\pm0.5}$ & $\mathbf{53.1\pm0.9}$ & $\mathbf{68.9\pm1.2}$\\

    \Xhline{2\arrayrulewidth}
    \end{tabular}
}
\end{table*}

\subsection{Comparisons on Zero-shot Recognition}
In the zero-shot setting, we evaluate our method across four video datasets using a pre-trained Kinetics-400 model, focusing on its generalization ability without fine-tuning. Results are presented under the full-class evaluation in Table \ref{tab:zero-shot}, demonstrating the effectiveness of our approach compared to state-of-the-arts based on CLIP image encoders as ViT-B/16 and ViT-L/14. This comparative analysis provides insights into the strengths of our spatio-temporal interaction module in video understanding tasks. 

Leveraging the ViT-B/16 CLIP image encoder, our method demonstrates superior performance compared to MAXI\cite{maxi} by 0.7\%, 0.4\%, and 0.5\% in top-1 accuracy on HMDB-51, UCF-101, and Kinetics-600, respectively. MAXI builds a text bag using captioner and GPT adopting video-specific attributes and utilizes it as additional semantic information. Our method outperforms MAXI while building attributes with less effort, which proves to be a significant result. 

Further tested with the ViT-L/14 CLIP image encoder, our approach outperforms BIKE\cite{bike} by 0.2\%, 0.3\%, and 0.4\% in top-1 accuracy on HMDB-51, UCF-101, and Kinetics-600, respectively. The BIKE model proposed a method to explore the bidirectional knowledge of video and text by using attributes to explore video-to-text directional knowledge and text-to-video directional knowledge through a temporal concept spotting mechanism. Our model is inspired by BIKE's temporal concept spotting mechanism, but we propose a spatio-temporal interaction module to learn attributes that are further relevant to videos and classes in patch-wise granularity. The results show that our method outperforms BIKE by adopting the proposed spatio-temporal interaction.

\begin{table*}[t]
\caption{Comparison of few-shot action recognition results. We present the performance with and without pretraining on Kinetics-400. The best results are highlighted in \textbf{bold}.}
\centering
\label{tab:few-shot}
\resizebox{1.0\linewidth}{!}{
    \begin{tabular}{lcccccccc}
    \Xhline{2\arrayrulewidth}
    \noalign{\vspace{3pt}}
    \multirow{2}{*}{\textbf{Method}} & \multicolumn{4}{c}{\textbf{UCF-101}} & \multicolumn{4}{c}{\textbf{HMDB-51}}\\
    \noalign{\vspace{1pt}}
    & $K=2$ & $K=4$ & $K=8$ & $K=16$ & $K=2$ & $K=4$ & $K=8$ & $K=16$\\
    \noalign{\vspace{1pt}}
    \Xhline{2\arrayrulewidth}
    \noalign{\vspace{2pt}}
    
    ActionCLIP\cite{actionclip} & 80.0 & 85.0 & 89.0 & - & 55.0 & 56.0 & 58.0 & - \\
    X-Florence\cite{x-clip} & 84.0 & 88.5 & 92.5 & 94.8 & 51.6 & 57.8 & 64.1 & 64.2 \\
    X-CLIP-B/16\cite{x-clip} & 80.0 & 85.0 & 89.0 & - & 55.0 & 56.0 & 58.0 & - \\
    MAXI (ViT-B/16)\cite{maxi} & 86.8 & 89.3 & 92.4 & 93.5 & 58.0 & 60.1 & 65.0 & 66.5\\
    ASU-B/16\cite{asu} & 91.4 & 94.6 & 96.0 & 97.2 & 60.1 & 63.8 & 67.2 & 70.8\\
    \noalign{\vspace{1pt}}

    \rowcolor{Gray}\textbf{Ours (ViT-B/16)} & \textbf{91.7} & \textbf{95.0} & \textbf{97.4} & \textbf{97.5} & \textbf{60.4} & \textbf{63.9} & \textbf{67.9} & \textbf{71.2}\\
    
    \Xhline{2\arrayrulewidth}
    \end{tabular}
}
\end{table*}

\subsection{Comparisons on Few-shot Recognition}
To assess the capability of capturing generalized representations from a minimal number of examples, we conduct few-shot recognition experiments using our method. We extend our model to classify all categories in the dataset using only $K$ samples per category for training. Results of $K$-shot learning are presented in Table \ref{tab:few-shot}. Our method achieves state-of-the-art performance compared to methods employing cross-modal pre-trained models, demonstrating robustness across different datasets and settings.

On UCF-101, our approach achieves a top-1 accuracy of 91.7\% under $K=2$, surpassing MAXI\cite{maxi} by 4.9\% and ASU by 0.3\%. For HMDB-51, our method achieves a top-1 accuracy of 60.4\% under $K=2$, outperforming ASU\cite{asu} by 0.3\%. Our method maintains state-of-the-art performance from $K=2$ to $K=16$, highlighting its effectiveness in leveraging limited examples for accurate classification tasks.

\begin{table*}[t]
\caption{Comparison with state-of-the-art methods on Kinetics400. \enquote{Views} indicates the number of temporal clips multiplied by the number of spatial crops during inference. The top performance is highlighted in \textbf{bold}.}
\centering
\label{tab:supervised}
\resizebox{1.0\linewidth}{!}{
    \begin{tabular}{lccccccc}
    \Xhline{2\arrayrulewidth}
    \noalign{\vspace{3pt}}
    \textbf{Method} & \textbf{Input} & \textbf{Pretrain} & \textbf{Top-1} & \textbf{Top-5} & \textbf{Views} & \textbf{Param(M)} & \textbf{FLOPs(G)}\\
    
    \Xhline{2\arrayrulewidth}
    \noalign{\vspace{1pt}}
    
    \multicolumn{8}{l}{\textit{Methods w/ large-scale image pre-training}} \\
    \noalign{\vspace{1pt}}
    ViViT-L/16\cite{vivit} & $32 \times 320^2$ & JFT-300M & 83.5 & 95.5 & $4 \times 3$ & 310.8 & 3992\\
    ViViT-H/16\cite{vivit} & $32 \times 224^2$ & JFT-300M & 84.8 & 95.8 & $4 \times 3$ & 647.5 & 8316\\
    TokenLearner\cite{tokenlearner} & $32 \times 224^2$ & JFT-300M & 85.4 & 96.3 & $4 \times 3$ & 450 & 4076\\ 
    MTV-H\cite{mtv} & $32 \times 224^2$ & JFT-300M & 85.8 & 96.6 & $4 \times 3$ & 450 & 3706\\ 
    CoVeR\cite{cover} & $16 \times 448^2$ & JFT-300M & 86.3 & - & $1 \times 3$ & - & -\\ 
    CoVeR\cite{cover} & $16 \times 448^2$ & JFT-3B & 87.2 & - & $1 \times 3$ & - & -\\ 
    \noalign{\vspace{2pt}}
    
    \hline
    \noalign{\vspace{1pt}}
    \multicolumn{8}{l}{\textit{Methods w/ large-scale vision-language pre-training}} \\
    \noalign{\vspace{1pt}}
    ActionCLIP (ViT-B/16)\cite{actionclip} & $16 \times 224^2$ & WIT-400M & 82.6 & 96.2 & $10 \times 3$ & 105.2 & 282\\
    ActionCLIP (ViT-B/16)\cite{actionclip} & $32 \times 224^2$ & WIT-400M & 83.8 & 97.1 & $10 \times 3$ & 141.7 & 563\\ 
    X-CLIP (ViT-L/14)\cite{x-clip} & $16 \times 336^2$ & WIT-400M & 87.7 & 97.4 & $4 \times 3$ & 451.2 & 3086\\ 
    Text4Vis (ViT-L/14)\cite{text4vis} & $32 \times 336^2$ & WIT-400M & 87.8 & 97.6 & $1 \times 3$ & 230.7 & 3829\\
    ASU (ViT-L/14)\cite{asu} & $16 \times 336^2$ & WIT-400M & 88.3 & 98.0 & $4 \times 3$ & 425.3 & 3084\\
    BIKE (ViT-L/14)\cite{bike} & $32 \times 336^2$ & WIT-400M & 88.6 & 98.3 & $4 \times 3$ & 230 & 3728\\
    
    \rowcolor{Gray}
    & $8 \times 224^2$ & WIT-400M & 87.3 & 97.7& $4 \times 3$ & 230 & 3740\\
    \rowcolor{Gray}
    & $16 \times 224^2$ & WIT-400M & 87.5 & 98.1 & $4 \times 3$ & 230 & 3740\\
    \rowcolor{Gray}
    \multirow{-3}{*}{\textbf{Ours (ViT-L/14)}} & $32 \times 336^2$ & WIT-400M & \textbf{88.8} & \textbf{98.6} & $4 \times 3$ & 230 & 3740\\

    \Xhline{2\arrayrulewidth}
    \end{tabular}
}
\end{table*}

\subsection{Comparisons on Fully-supervised Recognition}
In Table \ref{tab:supervised} , we present a detailed comparison with state-of-the-art methods on the Kinetics-400 dataset. Our method achieves outstanding results across different settings: at the $336\times336$ input resolution, utilizing 3 spatial crops and 4 temporal clips, we achieve a top-1 accuracy of 88.8\%. This performance surpasses the top-1 accuracy achieved by BIKE \cite{bike} by 0.3\%, establishing our method as a leader among CLIP-based approaches.

Furthermore, at the $224\times224$ input resolution setting with 16 frames, our approach achieves a top-1 accuracy of 87.5\%. This result represents a significant improvement over ActionCLIP \cite{actionclip} by 4.9\% under identical conditions. These findings highlight the robustness and effectiveness of our model in handling different input resolutions and frame settings, demonstrating its capability to achieve state-of-the-art performance in video classification tasks on Kinetics-400.

\begin{table*}[t]
\begin{multicols}{2}
\begin{table}[H]
    \caption{Ablation study on the effect of spatio-temporal interaction}
    \centering
    \label{tab:st-layer}
    \resizebox{1.\linewidth}{!}{
        \begin{tabular}{ccccc}
        \Xhline{2\arrayrulewidth}
        \noalign{\vspace{3pt}}
        \textbf{Spatial} & \textbf{Temporal} & \textbf{UCF101} & \textbf{HMDB51} & \textbf{K600}\\
        \noalign{\vspace{1pt}}
        \Xhline{2\arrayrulewidth}
        \noalign{\vspace{2pt}}
        - & - & 76.3 & 46.8 & 64.1\\
        \checkmark & - & 77.0 & 47.9 & 66.8\\
        - & \checkmark & 77.6 & 48.2 & 68.5\\
        \checkmark & \checkmark & \textbf{78.8} & \textbf{51.8} & \textbf{71.2}\\
        \noalign{\vspace{1pt}}
        \Xhline{2\arrayrulewidth}
        \end{tabular}
    }
\end{table}

\begin{table}[H]
    \caption{{Ablation study on different number of attributes}\vspace{2pt}}
    \centering
    \label{tab:num_attr}
    \resizebox{1.\linewidth}{!}{
        \begin{tabular}{cccc}
        \Xhline{2\arrayrulewidth}
        \noalign{\vspace{3pt}}
        \textbf{\#Attributes} & \textbf{UCF101} & \textbf{HMDB51} & \textbf{K600}\\
        \noalign{\vspace{1pt}}
        \Xhline{2\arrayrulewidth}
        \noalign{\vspace{2pt}}
        0 & 72.5 & 47.2 & 65.6\\
        2 & 78.2 & 51.3 & 70.5\\
        4 & 78.5 & 51.5 & 70.8\\
        8 & \textbf{78.8} & \textbf{51.8} & \textbf{71.2}\\
        16 & 78.6 & 51.7 & 70.9 \\
        \noalign{\vspace{1pt}}
        \Xhline{2\arrayrulewidth}
        \end{tabular}
    }
\end{table}
\end{multicols}
\end{table*}

\begin{table*}[t]
\begin{multicols}{2}
\begin{table}[H]
    \caption{{Ablation study on different fusion methods}\vspace{2pt}}
    \centering
    \label{tab:fusion}
    \resizebox{1.\linewidth}{!}{
        \begin{tabular}{lccc}
        \Xhline{2\arrayrulewidth}
        \noalign{\vspace{3pt}}
        \textbf{Fusion Method} & \textbf{Top-1 Acc} & \textbf{Top-5 Acc.} \\
        \noalign{\vspace{1pt}}
        \Xhline{2\arrayrulewidth}
        \noalign{\vspace{2pt}}
       
        Co-attention & 73.0 & 92.1 \\
        Co-attention + \textbf{ST} & 73.9 & 92.8 \\
        \textbf{ST (Ours)} & \textbf{78.9} & \textbf{94.5} \\
        \noalign{\vspace{1pt}}
        \Xhline{2\arrayrulewidth}
        \end{tabular}
    }
\end{table}
\end{multicols}
\end{table*}

\subsection{Ablation Study}
For all ablation experiments in this study, we employ ViT-B/16 with a configuration of 16 frames per video and conduct single-view inference. This setup ensures consistency across our evaluations, allowing us to systematically assess the impact of various modifications on model performance. We employ zero-shot inference methodology on the first split of the validation sets for UCF-101, HMDB-51, and Kinetics-600 datasets. This approach enables us to evaluate the generalization capability of our model without fine-tuning, providing insights into its effectiveness across different video datasets under controlled conditions.

\subsubsection{Effects of spatio-temporal interaction}
To examine the impact of spatio-temporal modules, we establish a baseline using ActionCLIP\cite{actionclip}, incorporating a 6-layer temporal transformer encoder atop the CLIP image encoder. Table \ref{tab:st-layer} presents the results, with the first row representing the baseline.
Compared to the baseline, integrating the spatial interaction module yields gains of 0.7\%, 1.1\%, and 2.7\% on UCF-101, HMDB-51, and Kinetics-600, respectively. Further, the addition of the temporal interaction module enhances accuracies by 1.3\%, 1.4\%, and 4.4\% on the respective datasets. Ultimately, combining spatio-temporal interaction achieves improvements of 2.5\%, 5.0\%, and 7.1\% on UCF-101, HMDB-51, and Kinetics-600. These results underscore the effectiveness of enhancing visual representations through spatio-temporal interaction.

\subsubsection{Effects of the number of description attributes}
In the study focusing on the number of description attributes as depicted in Table \ref{tab:num_attr}, we systematically vary the size of attributes, which corresponds to the number of keywords utilized. We observe a progressive enhancement in performance as we increase the attribute size from lower numbers to 8. This augmentation indicates that expanding the scope of descriptive attributes allows our model to capture more nuanced and specific characteristics, improving its ability to discern and classify video content accurately. Our primary experimental results, which demonstrate the effectiveness of our approach, are reported with an attribute size set at 8. This configuration not only optimizes performance but also ensures robustness across different video datasets.

\subsubsection{Effects of different fusion methods}
{We compare the effectiveness of various fusion strategies, including our proposed Spatio-Temporal Interaction module (denoted as ST). Specifically, we evaluate three different configurations: (1) using only Co-attention, (2) combining Co-attention with ST (Co-attention + ST), and (3) using only ST which is our proposed method. To isolate the impact of each fusion approach, all other network components and training settings are held constant. In Table} \ref{tab:fusion}, {simply applying co-attention provides a performance of 73.0\%. When we integrate our ST module (Co-attention + ST), we observe a +0.9\% gain in Top-1 accuracy compared to the co-attention only. Finally, combining ST with our baseline framework yields the highest accuracy, underlining the importance of a fine-grained spatio-temporal interaction alongside cross-attention.}

\begin{figure*}[t]
    \centering
    \includegraphics[width=1\linewidth]{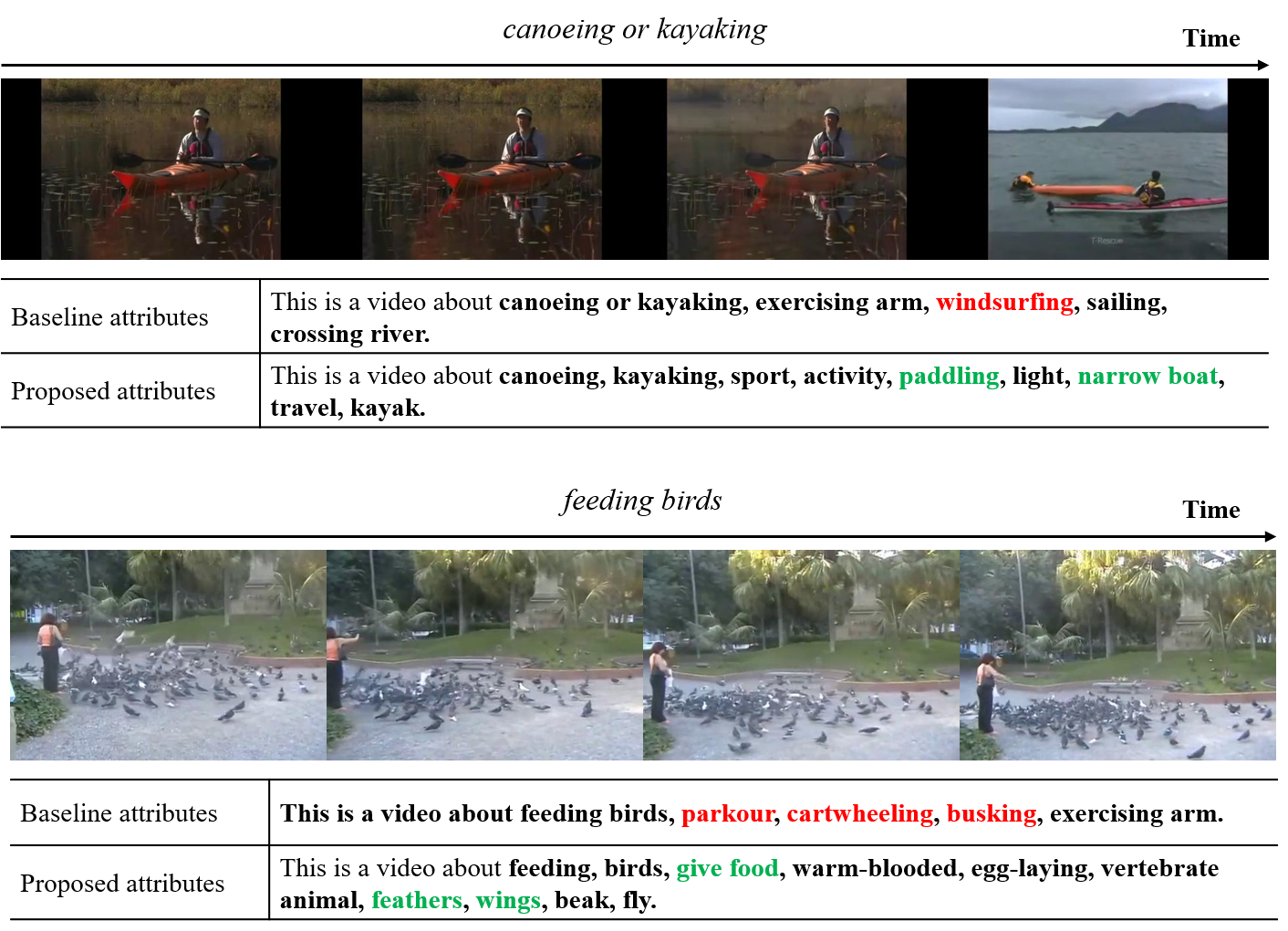}
    \caption{Qualitative comparison of attribute keywords. For two representative video samples, we show selected frames and attribute keywords produced by BIKE (baseline attributes) and by our description attributes (proposed attributes). Green text marks attributes that are closely aligned with the ground truth class, whereas red text marks attributes that are weakly related. Our method yields action and scene relevant descriptor compared to BIKE attributes.}
    \label{fig:visualization}
\end{figure*}

 \begin{figure*}[htbp]
    \centering
    \includegraphics[width=1\linewidth]{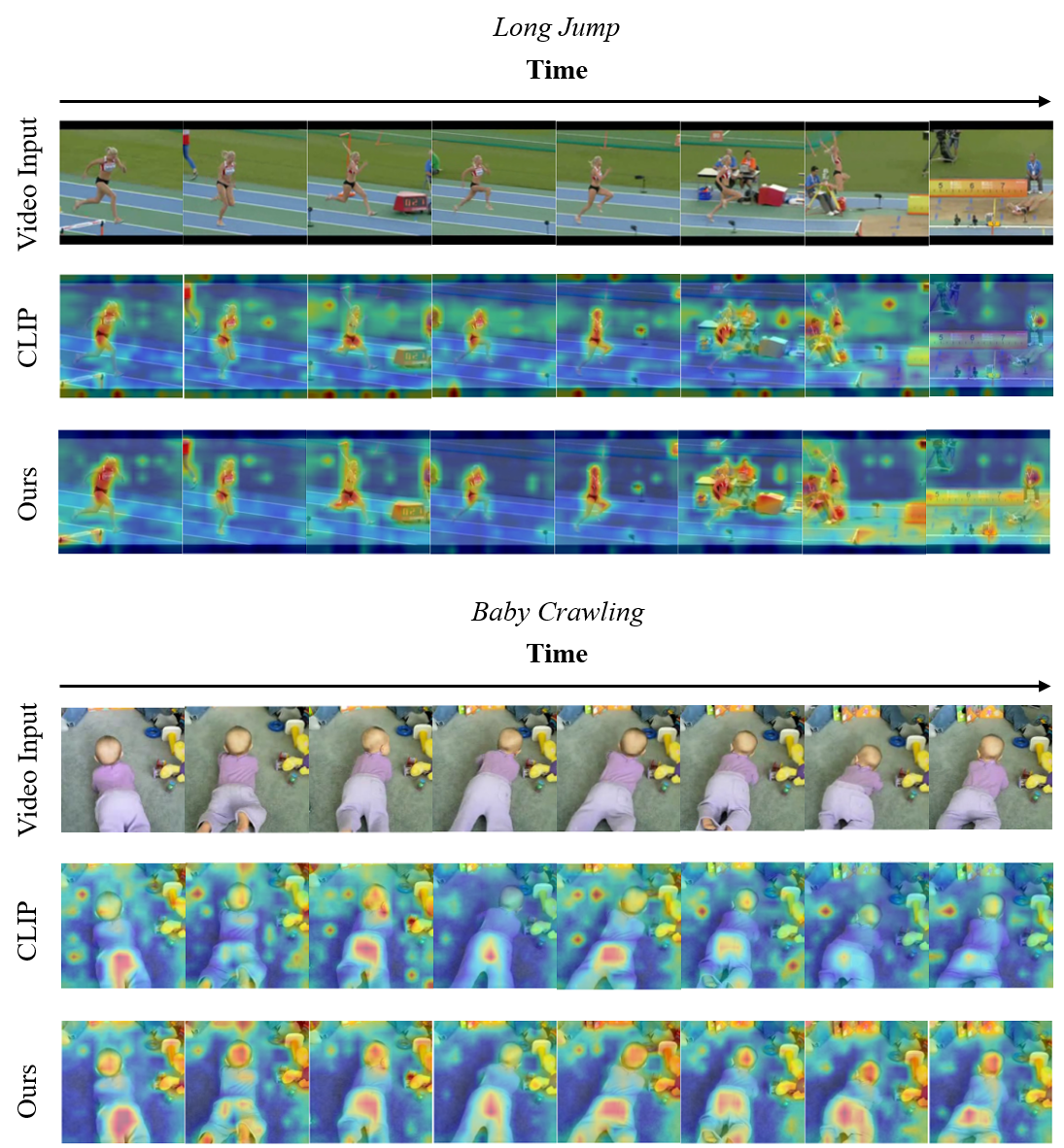}
    \caption{Attention map comparison on UCF-101 (zero-shot). For each sequence, the first row presents uniformly sampled frames, the second row shows attention from the CLIP baseline, and the third row shows attention produced by our spatio-temporal interaction guided by description attributes. Our method concentrates on action-relevant regions (e.g., athlete and landing area; baby torso/limbs) while suppressing background responses.}
    \label{fig:attention_map}
\end{figure*}

\subsection{Visualization}
\subsubsection{Comparative analysis of attribute relevance}
In Fig. \ref{fig:visualization}, we conduct a comparative analysis between the attributes generated by BIKE \cite{bike} and our proposed descriptive attributes. Upon closer examination of the attributes depicted in the figure, our descriptive attributes demonstrate superior relevance to the video content compared to the BIKE attributes, which sometimes include irrelevant descriptors. Our descriptive attributes delve into more intricate and specific dimensions, capturing nuanced details closely associated with the fine-grained dynamics of the depicted behavior in the video. This distinction highlights the effectiveness of our approach in refining attribute generation, thereby enhancing the model's ability to accurately interpret and classify video content based on meaningful and contextually relevant descriptors.

\subsubsection{Visualization of attention maps}
We visualize the attention maps to qualitatively evaluate the performance of our model in a zero-shot environment. We compared our method with vanila CLIP, examining how spatio-temporal module effects the model to directs its attention to different parts of the input data. This comparison allowed us to observe the distinctions in the focus patterns of both models. As shown in Fig. \ref{fig:attention_map}, our model demonstrates a stronger focus on highly relevant regions and a weaker focus on irrelevant regions compared to the vanila CLIP. This indicates that our model is more effective in identifying and concentrating on relevant areas within the content.
\medskip

\section{Conclusion}
In this paper, we introduce a novel approach utilizing description attributes composed of keywords extracted via a large language model. Our method harnesses action class descriptions to extract more meaningful words through a large-scale language model, enabling efficient and cost-effective zero-shot performance without reliance on video-specific attributes. Additionally, our spatio-temporal interaction module enhances alignment with descriptive attributes by considering spatial and temporal information at a fine-grained level, bridging the gap between attributes and the entire video embedding. Experimental results showcase the transferability of our model to downstream tasks, achieving top-1 accuracies of 81.0\%, 53.1\%, and 68.9\% on UCF-101, HMDB-51, and Kinetics-600, respectively.
As a future work, we plan to explore more advanced spatio-temporal modeling techniques to further enhance the alignment between video embeddings and descriptive attributes. We also intend to evaluate the scalability of our method on larger and more diverse datasets to validate its robustness and generalizability. We believe these directions will significantly contribute to the development of more comprehensive video understanding models.

\subsection{Limitations and Future Work}
{Despite demonstrating the effectiveness of descriptive attributes derived from a large language model and the proposed spatio-temporal interaction module, this work has several limitations. The quality of attributes highly depends on the reliability and diversity of external textual resources, potentially leading to less informative or irrelevant attributes. Another constraint is that extremely long video sequences or complex interactions may still pose a challenge in capturing fine-grained details. In future work, we plan to explore more robust attribute-generation methods, while also refining the spatio-temporal module to better handle complexities such as occlusion, motion blur, and multi-person activities.}

\section*{Acknowledgment}
This work was supported by Institute of Information \& communications Technology Planning \& Evaluation (IITP) grant funded by the Korea government(MSIT) (No. RS-2019-II190079, Artificial Intelligence Graduate School Program (Korea University), No. IITP-2025-RS-2025-02304828, Artificial Intelligence Star Fellowship support program to nurture the best talents, No. IITP-2025-RS- 2024-00436857, Information Technology Research Center, and No. RS-2024-00457882, AI Research Hub Project).
\medskip



 \bibliographystyle{elsarticle-num}
 \bibliography{reference.bib}





\end{document}